\documentclass[12pt, a4paper]{article}

\usepackage[latin2]{inputenc}
\usepackage{graphicx}
\usepackage{amsmath}
\usepackage{fancyhdr}
\usepackage{authblk}

\title{\bfseries Shannon entropy for intuitionistic fuzzy information}
\date{}
\author{\bfseries Vasile Patrascu \\ Research Center for Electronics and Information Technology,\\ Valahia University, Targoviste, Rom\^ania,\\email: patrascu.v@gmail.com }

\begin{document}

\maketitle
\thispagestyle{fancy}
\begin{abstract}
 The paper presents an extension of Shannon fuzzy entropy for intuitionistic fuzzy one. This extension uses a new formula for distance between two intuitionistic fuzzy pairs.\\

\end{abstract}

\section{Introduction}
The intuitionistic fuzzy representation of information was proposed by Atanassov [\ref{r1}], [\ref{r2}], [\ref{r3}] and it is defined by the pair  \(  \left(  \mu , \nu  \right)  \)  where  \(  \mu  \in  \left[ 0,1 \right]  \)  is the \textit{degree of truth} while  \(  \nu  \in  \left[ 0,1 \right]  \)  is the \textit{degree of falsity}. Also, Atanassov considered the following condition for the intuitionistic fuzzy pair  \(  \left(  \mu , \nu  \right)  \) :

\begin{equation}\label{1.1}
 \mu + \nu  \leq 1 
\end{equation}
 The condition (\ref{1.1}) permits to consider the third parameter, the\textit{ degree of incompleteness}  \(  \pi  \in  \left[ 0,1 \right]  \)  defined by:
 \begin{equation}\label{1.2}
\pi =1- \mu - \nu  \
\end{equation}
In addition to parameter  \(  \pi  \) , we define the\textit{ net truth}  \(  \tau \in  \left[ -1,1 \right]  \) , by:
\begin{equation}\label{1.3}
\tau= \mu - \nu
\end{equation}
On this way, we have two systems representation of intuitionistic fuzzy information: the primary space  \(  \left(  \mu , \nu  \right)  \)  or the explicit space and the secondary space  \(  \left(  \tau, \pi  \right)  \)  or the implicit space. Also, there exists a supplementary condition for the secondary space, namely:
\begin{equation}\label{1.4}
\vert  \tau \vert + \pi  \leq 1
\end{equation}
Taking into account the condition (\ref{1.4}), using the secondary space  \(  \left(  \tau, \pi  \right)  \)  we define the third parameter [\ref{r8}], \textit{degree of ambiguity}  \(  \alpha  \in  \left[ 0,1 \right]  \) , by:

\begin{equation}\label{1.5}
\alpha =1- \vert  \tau \vert - \pi
\end{equation}
The formulae (\ref{1.2}) and (\ref{1.3}) represent the transform from the primary space  \(  \left(  \mu , \nu  \right)  \)  to the secondary space  \(  \left(  \tau, \pi  \right)  \)  while the next two formulae represent the inverse transform:

\begin{equation}\label{1.6}
\mu =\frac{1- \pi + \tau}{2}
\end{equation}

\begin{equation}\label{1.7}
\nu =\frac{1- \pi - \tau}{2}
\end{equation}

For the intuitionistic\ fuzzy information   \( X= \left(  \mu , \nu  \right)  \) , it was defined \textit{the complement}  \(\bar{X}\)   by: 

\begin{equation}\label{1.8}
\bar{X}= \left(  \nu , \mu  \right)
\end{equation}

After presentation of the main parameters that will be used in this approach, the next will have the following structure: section two presents a new distance for intuitionistic fuzzy information; section three presents formulae for evaluating of some feature of intuitionistic fuzzy information like \textit{certainty}, \textit{score}, \textit{uncertainty}; section four presents  the \textit{escort fuzzy information}; section five presents the \textit{Shannon entropy}  formula for intuitionistic fuzzy information; section six presents the conclusion while the last is the references section.

\section{A distance for intuitionistic fuzzy information}

In this section we define a new distance for intuitionistic fuzzy pairs. For two intuitionistic fuzzy pairs  \( P= \left(  \mu _{p}, \nu _{p} \right)  \)  and  \( Q= \left(  \mu _{q}, \nu _{q} \right)  \) , we consider the  \( L1 \)  distance  \( d \left( P,Q \right)  \in  \left[ 0,2 \right]  \)  define by:

\begin{equation}\label{2.1}
 d \left( P,Q \right) = \vert  \mu _{p}- \mu _{q} \vert + \vert  \nu _{p}- \nu _{q} \vert 
\end{equation}

The  \( L1 \)  distance [\ref{r10}], [\ref{r11}] is a metric and considering the auxiliary point  \( C= \left( 1,1 \right)  \)  (see Figure 1), there exists the triangle inequality, namely:

\begin{equation}\label{2.2}
d \left( P,C \right) +d \left( C,Q \right)  \geq d \left( P,Q \right)
\end{equation}
Because 

\begin{equation}\label{2.2a}
 d \left( P,C \right) +d \left( C,Q \right)  \neq 0
\end{equation}

we can transform (\ref{2.2}) into (\ref{2.3}):

\begin{equation}\label{2.3}
1 \geq \frac{d \left( P,Q \right) }{d \left( P,C \right) +d \left( C,Q \right) }
\end{equation}

The right term represents the new distance or the new dissimilarity, namely:

\begin{equation}\label{2.4}
D \left( P,Q \right) =\frac{d \left( P,Q \right) }{d \left( P,C \right) +d \left( C,Q \right) }
\end{equation}

\begin{figure}[hhp]\label{figure1}
	\centering
		\includegraphics[width=2.91in,height=2.35in]{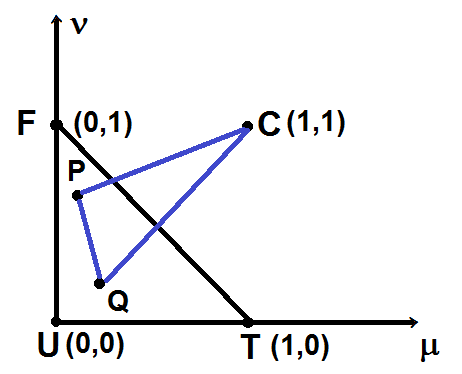}
		\caption{The geometrical framework for the proposed intuitionistic fuzzy distance.}
\end{figure}

From (\ref{2.4}) and (\ref{2.1}) it results the distance formula:
\begin{equation}\label{2.5}
D \left( P,Q \right) =\frac{ \vert  \mu _{p}- \mu _{q} \vert + \vert  \nu _{p}- \nu _{q} \vert }{2+ \pi _{p}+ \pi _{q}}
\end{equation}
Using the clasical negation it results the following similarity formula:
\begin{equation}\label{2.6}
S \left( P,Q \right) =1-\frac{ \vert  \mu _{p}- \mu _{q} \vert + \vert  \nu _{p}- \nu _{q} \vert }{2+ \pi _{p}+ \pi _{q}}
\end{equation}
We notice that distance (\ref{2.5}) and its similarity (\ref{2.6}) take values in the interval  \(  \left[ 0,1 \right]  \). Also, this distance is not a metric because it does not verify the triangle inequality.

\section{The certainty, the score and the uncertainty for intuitionistic fuzzy information}

Starting from the proposed distance defined by (\ref{2.5}), we will construct some measures for the following three features of intuitionistic fuzzy information: the certainty, the score and the uncertainty.

\subsection{The intuitionistic fuzzy certainty}
For any intuitionistic fuzzy pair  \( X= (  \mu , \nu )  \)  we consider its complement  \(\bar{ X}= (  \nu , \mu )  \)  and we define the \textit{certainty} as dissimilarity between  \( X \)  and  \(\bar{ X} \) , namely:
\begin{equation}\label{3.1}
g( X ) =D \left( X,\bar{X} \right)
\end{equation}

with its equivalent form:

\begin{equation}\label{3.2}
g(X)=\frac{\vert \mu -\nu \vert }{2-\mu -\nu }
\end{equation}

In the space $(\mu ,\nu )$ we identify the following properties for \textit{intuitionistic fuzzy  certainty}: 

\newcounter{numberedCntDI}
\begin{enumerate}
\item $g(1,0)=g(0,1)=1$
\item $g(x,x)=0$
\item $g(\mu ,\nu )=g(\nu ,\mu )$
\item $g(\mu _{1},\nu _{1})\le g(\mu _{2},\nu _{2})$ if $\vert \mu 
_{1}-\nu _{1}\vert \le \vert \mu _{2}-\nu _{2}\vert $ and $\mu _{1}+\nu 
_{1}\le \mu _{2}+\nu _{2}$
\setcounter{numberedCntDI}{\theenumi}
\end{enumerate}

The property (4) shows that the certainty increases with $\vert \tau \vert $ and decreases with $\pi $.

From property (4) it results that $g(\mu ,\nu ) \in [0,1]$ because $g(\mu ,\nu 
)\ge g(0,0)$ and $g(\mu ,\nu )\le g(1,0)$.

\subsection{The intuitionistic fuzzy score}

From (\ref{3.2}) came the idea to define the \textit{intuitionistic fuzzy score} by:
\begin{equation}\label{3.3}
r(X)=\frac{\mu -\nu }{2-\mu -\nu }
\end{equation}

with the equivalent form in the space $(\tau ,\pi )$, [\ref{r6}]:
\begin{equation}\label{3.4}
r(X)=\frac{\tau }{1+\pi } 
\end{equation}

In the space $(\mu ,\nu )$ the properties for the \textit{intuitionistic fuzzy score} derive from the certainty properties, namely:

\newcounter{numberedCntDJ}
\begin{enumerate}
\item $r(1,0)=1;r(0,1)=-1$
\item $r(x,x)=0$
\item $r(\mu ,\nu )=-r(\nu ,\mu )$
\item $r(\mu _{1},\nu _{1})\le r(\mu _{2},\nu _{2})$ if $\mu _{1}-\nu 
_{1}\le \mu _{2}-\nu _{2}$ and $\mu _{1}+\nu _{1}\le \mu _{2}+\nu _{2}$
\setcounter{numberedCntDJ}{\theenumi}
\end{enumerate}

The property (4) shows that the \textit{intuitionistic fuzzy score} 
increases with $\tau $ and decreases with $\pi $. From property (4) it 
results that $r(\mu ,\nu ) \in [-1,1]$ because $r(\mu ,\nu )\ge r(0,1)$ 
and $r(\mu ,\nu )\le r(1,0)$.

\subsection{The intuitionistic fuzzy uncertainty}

Finally, we define the \textit{intuitionistic fuzzy} \textit{
uncertainty} using the negation of the certainty:
\begin{equation}\label{3.5}
 e(X)=1-\frac{\vert \mu -\nu \vert }{2-\mu -\nu } 
\end{equation}

with the equivalent form in the space $(\tau ,\pi )$ $[7]$:
\begin{equation}\label{3.6}
e(X)=1-\frac{\vert \tau \vert }{1+\pi } 
\end{equation}

In the space $(\mu ,\nu )$ the \textit{intuitionistic fuzzy  uncertainty} verifies the following conditions [\ref{r6}]:

\newcounter{numberedCntEA}
\begin{enumerate}
\item $e(1,0)=e(0,1)=0$
\item $e(x,x)=1$
\item $e(\mu ,\nu )=e(\nu ,\mu )$
\item $e(\mu _{1},\nu _{1})\le e(\mu _{2},\nu _{2})$ if $\vert \mu 
_{1}-\nu _{1}\vert \ge \vert \mu _{2}-\nu _{2}\vert $ and $\mu _{1}+\nu 
_{1}\ge \mu _{2}+\nu _{2}$
\setcounter{numberedCntEA}{\theenumi}
\end{enumerate}

The property (4) shows that the uncertainty decreases with $\vert \tau \vert $ and increases with $\pi $.

From property (4) it results that $e(\mu ,\nu )\in[0,1]$ because $e(\mu ,\nu 
)\ge e(1,0)$ and $e(\mu ,\nu )\le e(0,0)$

\section{The escort fuzzy information}

We will associate to any intuitionistic fuzzy information $X=(\mu ,\nu 
)$ a fuzzy one $\hat{X}=(\hat{\mu },\hat{\nu })$ that we call \textit{escort 
fuzzy information}. The escort fuzzy pair $(\hat{\mu },\hat{\nu })$ will be 
determined in order to preserve the score of the intuitionistic fuzzy 
pair $(\mu ,\nu )$. It will be obtained by solving the following 
system:
\begin{equation}\label{4.1}
\hat{\mu }+\hat{\nu }=1
\end{equation}
\begin{equation}\label{4.2}
\hat{\mu }-\hat{\nu }=r(\mu ,\nu )
\end{equation}

It results the following values for the escort fuzzy pair ($\hat{\mu },\hat{\nu })$: 

\begin{equation}\label{4.3}
\hat {\mu }=\frac{\mu +\pi }{1+\pi }
\end{equation}

\begin{equation}\label{4.4}
\hat {\nu }=\frac{\nu +\pi }{1+\pi }
\end{equation}

There exist the following two inequalities:
\begin{equation}\label{4.5}
\mu +\pi \ge \hat{\mu }\ge \mu
\end{equation}

\begin{equation}\label{4.6}
\nu +\pi \ge \hat{\nu }\ge \nu
\end{equation}

The escort fuzzy pair $(\hat{\mu },\hat{\nu })$ can be used to extend existing 
results from fuzzy theory [\ref{r12}], [\ref{r13}] to intuitionistic fuzzy one. 
An example could be the cardinal's calculation of an intuitionistic 
fuzzy set using its escort fuzzy set [\ref{r6}]. In this paper we will use 
the escort pair for extending the Shannon fuzzy entropy to Shannon 
intuitionistic fuzzy entropy.

\section{The Shannon entropy for intuitionistic fuzzy information}

For any intuitionistic fuzzy pair $X=(\mu ,\nu )$, using the escort 
fuzzy pair $\hat{X}=(\hat{\mu },\hat{\nu })$ we will define the Shannon entropy [\ref{r9}],  [\ref{r5}] by the formula:
\begin{equation}\label{5.1}
 E_{S}(X)= e_{S}(\hat{X}) 
\end{equation}
where  $E_{S}$ represents the Shannon entropy for intuitionistic fuzzy information while $e_{S}$ represents the Shannon entropy for fuzzy one [\ref{r5}].\\
For fuzzy information, De Luca and Termini [\ref{r5}] extended the Shannon 
formula for calculating the fuzzy entropy by:
\begin{equation}\label{5.2}
 e_{S}(\mu )=-\mu \ln (\mu )-(1-\mu )\ln (1-\mu )
\end{equation}

From (\ref{5.2}) it results:
\begin{equation}\label{5.3}
e_{S}(\hat{X})=-\hat{\mu }\ln (\hat{\mu })-\hat{\nu }\ln (\hat{\nu })
\end{equation}

From (\ref{4.3}), (\ref{4.4}), (\ref{5.1}) and (\ref{5.3}) it results the Shannon variant for 
intuitionistic fuzzy entropy:

\begin{equation}\label{5.4}
E_{S}(X)=-\frac{\mu +\pi }{1+\pi } \ln \left( \frac{\mu +\pi }{1+\pi 
}\right ) -\frac{\nu +\pi }{1+\pi }\ln \left(\frac{\nu +\pi }{1+\pi }\right)
\end{equation}

There are the next four equivalent formulae:

Using (\ref{1.6}) and (\ref{1.7}) it results:

\begin{equation}\label{5.5}
E_{S}(X)=-\frac{1+\dfrac{\tau}{1+\pi }}{2}\ln \left( \frac{1+\dfrac{\tau}{1+\pi }}{2} \right)-\frac{1-\dfrac{\tau}{1+\pi }}{2}\ln \left( \frac{1-\dfrac{\tau}{1+\pi }}{2} \right)
\end{equation}

Using (\ref{3.4}) it results:

\begin{equation}\label{5.6}
 E_{S}(X)=-\frac{1+r}{2}\ln \left(\frac{1+r}{2}\right)-\frac{1-r}{2}\ln 
\left(\frac{1-r}{2}\right)
\end{equation}

Because in (\ref{5.6}) there exists symmetry between $r$ and $-r$, it results:

\begin{equation}\label{5.7}
E_{S}(X)=-\frac{1+\vert r\vert }{2}\ln \left(\frac{1+\vert r\vert}{2}\right)-\frac{1-\vert r\vert }{2}\ln \left(\frac{1-\vert r\vert }{2}\right)
\end{equation}

From (\ref{3.2}), (\ref{3.3}) and (\ref{5.7}) it results:

\begin{equation}\label{5.8}
E_{S}(X)=-\frac{1+g}{2}\ln \left(\frac{1+g}{2}\right)-\frac{1-g}{2}\ln \left(\frac{1-g}{2}\right) 
\end{equation}

We notice that:

\begin{equation}\label{5.9}
\frac{\partial g}{\partial \vert \tau \vert }=\frac{1}{1+\pi }
\end{equation}

\begin{equation}\label{5.10}
 \frac{\partial g}{\partial \pi }=-\frac{\vert \tau \vert }{(1+\pi 
)^{2}}
\end{equation}

\begin{equation}\label{5.11}
 \frac{\partial E_{S}}{\partial g}=\frac{1}{2}\ln \left(\frac{1-g}{1+g}\right) 
\end{equation}

\begin{equation}\label{5.12}
 \frac{\partial E_{S}}{\partial \vert \tau \vert 
}=\frac{1}{2}\frac{1}{1+\pi }\ln \left(\frac{1-g}{1+g}\right)
\end{equation}

\begin{equation}\label{5.13}
 \frac{\partial E_{S}}{\partial \pi }=-\frac{1}{2}\frac{\vert \tau 
\vert }{(1+\pi )^{2}}\ln \left(\frac{1-g}{1+g}\right) 
\end{equation}

Because

\begin{equation}\label{5.13a}
\frac{1-g}{1+g}\le 1
\end{equation}

it results:

\begin{equation}\label{5.14}
\ln \left(\frac{1-g}{1+g}\right)\le 0 
\end{equation}

From (\ref{5.12}), (\ref{5.13}) and (\ref{5.14}) it results:

\begin{equation}\label{5.15}
 \frac{\partial E_{S}}{\partial \vert \tau \vert }\le 0
\end{equation}

\begin{equation}\label{5.16}
\frac{\partial E_{S}}{\partial \pi }\ge 0 
\end{equation}

As conclusion, it results that the Shannon entropy for intuitionistic 
fuzzy information defined by (\ref{5.4}) verifies the condition (4) from 
subsection 3.3, namely it decreases with $\vert \tau \vert $ and increases 
with $\pi $.

Also the function $E_{S}$ defined by (\ref{5.4}) verifies the conditions (1) 
and (3). In order to verify the condition (2) it necessary to multiply by the well-known normalization factor:

\begin{equation}\label{5.17}
 \lambda =\frac{1}{ln(2)} 
\end{equation}

Finally it results the normalized variant for Shannon entropy, namely:
\begin{equation}\label{5.18}
 E_{SN}(X)=-\frac{1}{\ln (2)} \left[ \frac{\mu +\pi }{1+\pi }\ln \left( \frac{\mu 
+\pi }{1+\pi }\right)+\frac{\nu +\pi }{1+\pi }\ln \left(\frac{\nu +\pi }{1+\pi }  \right)\right] 
\end{equation}

We can decompose the normalized Shannon entropy $E_{SN}$ in a sum with two terms, fuzziness $E_{A}$ and incompleteness $E_{U}$, namely:

\begin{equation}\label{5.19}
 E_{SN}(X)=E_{A}(X)+E_{U}(X) 
\end{equation}

where

\begin{equation}\label{5.20}
 E_{A}(X)=-\frac{(\mu +\pi )\ln (\mu +\pi )+(\nu +\pi )\ln (\nu +\pi )}{(1+\pi )\ln (2)} 
\end{equation}

\begin{equation}\label{5.21}
 E_{U}(X)=\frac{\ln (1+\pi )}{\ln (2)}
\end{equation}

From Jensen inequality [\ref{r4}] it results that:

\begin{equation}\label{5.22}
 E_{A}(X)\le -\frac{1}{\ln (2)}\ln \left(\frac{1+\pi }{2}\right)
\end{equation}

From (\ref{5.22}) it results that $E_{A}(X)$ is maximum for $X=(0.5,0.5)$ 
while from (\ref{5.21}) it results that $E_{U}(X)$ is maximum for $X=(0,0)$. Thus, it is highlighted that the uncertainty of intuitionistic fuzzy information has two sources: fuzziness (or ambiguity) that is the similarity of the pairs $(\mu ,\nu )$ with $(0.5,0.5)$ and incompleteness (or ignorance) that is the similarity of the pairs $(\mu ,\nu )$ with $(0,0).$

\section{Conclusion}

In this paper, we presented a new formula for calculating the distance and similarity of intuitionistic fuzzy information. Then, we constructed measures for information features like score, certainty and uncertainty. Also, a new concept was introduced, namely escort fuzzy information. 
Then, using the escort fuzzy information, Shannon's formula for intuitionistic fuzzy information was obtained. It should be underlined that Shannon's entropy for intuitionistic fuzzy information verifies the four defining conditions of intuitionistic fuzzy uncertainty. The measures of its two components were also identified: fuzziness (ambiguity) and incompleteness (ignorance).

\section*{References}

\begin{enumerate}
\item\label{r1} Atanassov, K. (1986) Intutitionistic fuzzy sets, \textit{Fuzzy Sets Syst.}, 20, 87-96.
\item\label{r2}Atanassov, K. (1999) \textit{Intuitionistic Fuzzy Sets: Theory 
and Applications. Studies in Fuzziness and Soft Computing}, vol 35, Physica-Verlag, Heidelberg.
\item\label{r3} Atanassov, K. (2012) \textit{On Intuitionistic Fuzzy Sets Theory }, Springer, Berlin.
\item\label{r4} Jensen, J.L.W.V. (1906) Sur les fonctions convexes et les 
inegalites entre les valeurs moyennes, \textit{Acta Mathematica}. 30(1), 175-193.
\item\label{r5} De Luca, A., Termini, S. (1972) A definition of nonprobabilistic 
entropy in the setting of fuzzy theory. \textit{Information and Control 20}, 301-312.
\item\label{r6} Patrascu, V. (2010) Cardinality and Entropy for Bifuzzy Sets, 
\textit{Proceedings of the 13$^{th}$ International Conference on 
Information Processing and Management of Uncertainty IPMU 2010}, 28 
Jun-02 Jul 2010, Dortmund, Germany, Part I, CCIS 80, 656-665.
\item\label{r7} Patrascu, V. (2012) Fuzzy membership function construction based 
on multi-valued evaluation, \textit{Proceedings of 10$^{th}$ 
International Conference FLINS}, 26-29 August 20102, Istanbul, Turkey, 
Uncertainty Modeling in Knowledge Engineering and Decision Making, 756-761.
\item\label{r8} Patrascu, V. (2016) Refined Neutrosophic Information Based on 
Truth, Falsity, Ignorance, Contradiction and Hesitation, \textit{Neutrosophic Sets and Systems}, Vol. 11, 57-66.
\item\label{r9} Shannon, C. E. (1948) A mathematical theory of communication, 
\textit{Bell System Tech}., J. 27, 379-423.
\item\label{r10} Website.com,\textit{ http://en.wikipedia.org/wiki/Lp\_space}
\item\label{r11} Website.com,\textit{ http://en.wikipedia.org/wiki/Taxicab\_geometry}
\item\label{r12} Zadeh, L. (1965) Fuzzy sets, \textit{Inform and Control},8, 338-353.
\item\label{r13}Zadeh, L. (1965) Fuzzy sets and systems, \textit{Proc. Symp. On 
Systems Theory}, Polytechnic Institute of Brooklyn, New York, 29-37.

\end{enumerate}

\end{document}